# Enhanced Boolean Correlation Matrix Memory

Mario Mastriani

*Abstract*— This paper introduces an Enhanced Boolean version of the Correlation Matrix Memory (CMM), which is useful to work with binary memories. A novel Boolean Orthonormalization Process (BOP) is presented to convert a non-orthonormal Boolean basis, i.e., a set of non-orthonormal binary vectors (in a Boolean sense) to an orthonormal Boolean basis, i.e., a set of orthonormal binary vectors (in a Boolean sense). This work shows that it is possible to improve the performance of Boolean CMM thanks BOP algorithm. Besides, the BOP algorithm has a lot of additional fields of applications, e.g.: Steganography, Hopfield Networks, Bi-level image processing, etc. Finally, it is important to mention that the BOP is an extremely stable and fast algorithm.

## I. INTRODUCTION

AN associative memory may also be classified as linear or nonlinear, depending on the model adopted for its neurons [1]. In the linear case, the neurons act (to a first approximation) like a linear combiner [2-6]. To be more specific, let the data vectors **a** and **b** denote the stimulus (input) and the response (output) of an associative memory, respectively. In a linear associative memory, the input-out-put relationship is described by

$$\mathbf{b} = \mathbf{M}\,\mathbf{a} \qquad (1)$$

where **M** is called the *memory matrix*. The matrix **M** specifies the network connectivity of the associative memory. Fig.1 depicts a block-diagram representation of a linear associative memory. In a nonlinear associative memory, on the other hand, we have an input-output relationship of the form

$$\mathbf{b} = \varphi\,(\mathbf{M};\mathbf{a})\,\mathbf{a} \qquad (2)$$

where, in general, φ(.;.) is a nonlinear function of the memory matrix and the input vector.

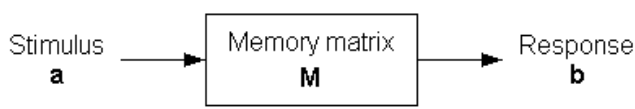

Fig.1 Block diagram of associative memory

Although this form of matrix was introduced in the 1960's and has been studied intensively since then, see [7-11], and the use of orthogonalization is not new to improve the storage of these networks, see [9, 10], unfortunately, a CMM version doesn't exist when the *key pattern* $\mathbf{a}_k$ and the *memorized pattern* $\mathbf{b}_k$ for all *k* are binary.

## II. METHODS

### A. Orthogonality in a Boolean sense

Given a set of binary vectors $\mathbf{u}_k = [\ u_{k1}\ ,\ u_{k2}\ ,\ \ldots\ ,\ u_{kp}\ ]^T$ (where $k = 1, 2, \ldots, q$, and $[.]^T$ means transpose of $[.]$), *they are orthonormals in a Boolean sense*, if they satisfy the following pair of conditions:

$$\mathbf{u}_k^T \wedge \mathbf{u}_j = 1 \text{ if } k = j \qquad (3.1)$$

and

$$\mathbf{u}_k^T \wedge \mathbf{u}_j = 0 \text{ if } k \neq j \qquad (3.2)$$

where the term $\mathbf{u}_k^T \wedge \mathbf{u}_j$ represents the *inner AND operation between binary vectors* $\mathbf{u}_k$ and $\mathbf{u}_j$, i.e.,

$$\mathbf{u}_k^T \wedge \mathbf{u}_j = (u_{k1} \wedge u_{j1}) \vee (u_{k2} \wedge u_{j2}) \vee \bullet\bullet\bullet \vee (u_{kp} \wedge u_{jp})$$

$$= \bigvee_{n=1}^{p} (\mathbf{u}_{kn} \wedge \mathbf{u}_{jn}) \qquad (4)$$

where ∨ represents the OR operation.

### B. Boolean Correlation Matrix Memory

Suppose that an associative memory has learned the memory matrix **M** through the associations of binary key and memorized patterns described by $\mathbf{a}_k \rightarrow \mathbf{b}_k$, where $k = 1, 2, \ldots, q$, being $\mathbf{a}_k$ a *p-by-1* binary key pattern vector, i.e., $\mathbf{a}_k = [\ a_{k1}, a_{k2}, \ldots, a_{kp}\ ]^T$, and $\mathbf{b}_k$ a *p-by-1* binary memorized pattern vector, i.e., $b_k = [\ b_{k1}, b_{k2}, \ldots, b_{kp}\ ]^T$, i.e., whose values are exclusively 0 or 1.

We may postulate *M*, denoting an *estimate* of the memory matrix **M** in terms of these patterns as follows

$$M = \bigvee_{k=1}^{q} (\mathbf{b}_k \wedge \mathbf{a}_k^T) \qquad (5)$$

where ∧ represents the AND operation, and the term $\mathbf{b}_k \wedge \mathbf{a}_k^T$ represents the *outer AND operation between binary vectors* of the key pattern $\mathbf{a}_k$ and and the memorized pattern $\mathbf{b}_k$.

While, in MATLAB code:

```
M = zeros(p,p);
for k = 1:q
   M = M | (b(:,k) & a(:,k)');
end
```

Therefore, *M* is a *p-by-p* binary estimate memory matrix. Where | represents the OR operation, and & represents the AND operation in MATLAB code, see Fig.2.

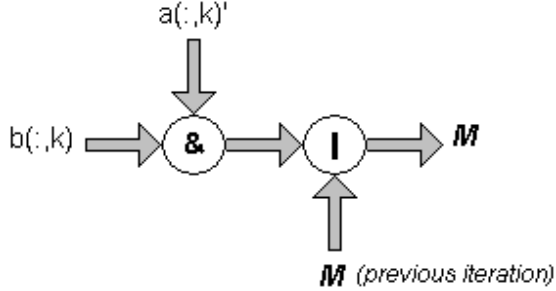

Fig.2 Signal-flow graph representation of Boolean CMM in MATLAB syntax

*C. Recall*

The fundamental problem posed by the use of an Boolean Associative Memory (BAM) is the address and re-call of patterns stored in memory, which is similar to non-Boolean associative memory [1]. To explain one aspect of this problem, let *M* denote the memory matrix of a BAM, which has been completely learned through its exposure to *q* pattern associations in accordance with Eq.5. Let a key pattern $\mathbf{a}_j$ be picked at random and reapplied as *stimulus* to the memory, yielding the *response*

$$\mathbf{b} = M \wedge \mathbf{a}_j \qquad (6)$$

The term $M \wedge \mathbf{a}_j$ represents the *AND operation* between binary memory matrix *M* and the key pattern $\mathbf{a}_j$. Substituting Eq.5 in 6, we get

$$\mathbf{b} = \bigvee_{k=1}^{q} (\mathbf{b}_k \wedge \mathbf{a}_k^T) \wedge \mathbf{a}_j$$
$$= \bigvee_{k=1}^{q} (\mathbf{a}_k \wedge \mathbf{a}_k^T) \wedge \mathbf{b}_k \qquad (7)$$

where, in the second line, it is recognized that $\mathbf{a}_k^T \wedge \mathbf{a}_j$ is a Boolean element equal to the *inner AND operation between binary vectors* of the key vectors $\mathbf{a}_k$ and $\mathbf{a}_j$. Moreover, we may rewrite Eq.7 as

$$\mathbf{b} = \left( (\mathbf{a}_j^T \wedge \mathbf{a}_j) \wedge \mathbf{b}_j \right) \vee \left( \bigvee_{(k=1) \wedge (k \neq j)}^{q} (\mathbf{a}_k^T \wedge \mathbf{a}_j) \wedge \mathbf{b}_k \right) \qquad (8)$$

We now see that if the key vectors have *Boolean orthonormality* (i.e., perpendicular to each other in a Boolean sense), then the response **b** equals $\mathbf{b}_j$. Accordingly, we may state that the Boolean memory associates perfectly if the key vectors form an *orthonormal set* (in a Boolean sense); that is, they satisfy the following pair of conditions:

$$\mathbf{a}_k^T \wedge \mathbf{a}_j = 1 \text{ if } k = j \qquad (9.1)$$

and

$$\mathbf{a}_k^T \wedge \mathbf{a}_j = 0 \text{ if } k \neq j \qquad (9.2)$$

Suppose now that the key vectors do form an orthonormal set (in a Boolean sense), as prescribed in Eq.9. What is then the limit on the *storage capacity* of the BAM ? Stated another way, what is the largest number of patterns that can be reliably stored ? The answer to this fundamental question lies the rank of the Boolean Memory Matrix (BMM) *M*. The rank of a matrix is defined as the number of independent columns (rows) of the matrix, being *rank* $\leq$ *min(p,p)*. The BMM *M* is a *p-by-p* matrix, where *p* is the dimensionality of the input space. Hence the rank of the BMM *M* is limited by the dimensionnality *p*. We may thus formally state that the number of patterns that can be reliably stored can never exceed the input space dimensionality.

*D. Boolean Orthonormalisation Process (BOP)*

Given a set of key binary vectors that are nonorthonormal (in a Boolean sense), we may use a *preprocessor* to transform them into an orthonormal set (in a Boolean sense); the preprocessor is designed to perform a *Boolean orthonorma-lization* on the key binary vectors prior to association. This form of transformation is described below, maintaining a one-to-one correspondence between the input (key) binary vectors $\mathbf{a}_1$, $\mathbf{a}_2$, ... , $\mathbf{a}_q$ and the resulting orthonormal binary vectors $\mathbf{c}_1$, $\mathbf{c}_2$, ... , $\mathbf{c}_q$, as indicated here:

$$\{ \mathbf{a}_1, \mathbf{a}_2, ... , \mathbf{a}_k \} \leftrightarrow \{ \mathbf{c}_1, \mathbf{c}_2, ... , \mathbf{c}_k \}$$

where $\mathbf{c}_1 = \mathbf{a}_1$, and the remaining $\mathbf{c}_k$ are defined by

$$\mathbf{c}_{j,k} = \mathbf{a}_{j,k} \veebar \left( \bigvee_{i=1}^{k-1} (\mathbf{c}_{j,i} \wedge \mathbf{a}_{j,k}) \right) \qquad (10)$$

with $k = 2, 3, ... , q$ and $j = 1, 2, ... , p$, and where $\veebar$ represents the XOR operation.

While, in MATLAB code:

*First version (by element)*

```
c(1,:) = a(1,:);
for j = 1:p
  for k = 2:q
    acu(j,k) = 0;
    for i = 1:k-1
      acu(j,k)= acu(j,k)|(c(j,i)&a(j,k));
    end
    c(j,k) = xor(a(j,k),acu(j,k));
  end
end
```

*Second version (by vector)*

```
c(:,1) = a(:,1);
for k = 2:q
  acu(:,k) = zeros(p,1);
  for i = 1:k-1
    acu(:,k) = acu(:,k)|(c(:,i)&a(:,k));
  end
  c(:,k) = xor(a(:,k),acu(:,k));
end
```

The associations are then performed on the pairs ($c_k$, $b_k$), $k = 2, 3, \ldots, q$. The block diagram Fig.3 highlights the order in which the preprocessing and association are performed.

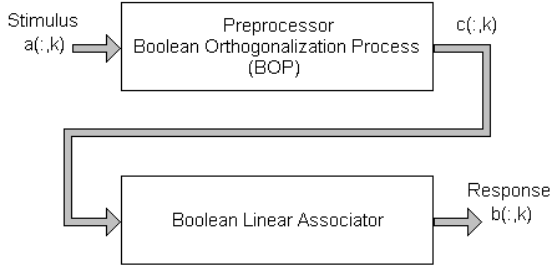

Fig.3 Extension of Boolean associative memory in MATLAB syntax

*E. Simulations Results*

**Simulation of BOP Algorithm**

Based on Fig.4, we have 6 binary vectors (columns) of 7 elements each one, called a(:,k) set. Applying the BOP algorithm to this set, another set of binary vectors is obtained, called c(:,k) but whose vectors are orthonormal to each other. Observe that to the right of the first **1** of each row of c(:,k) set there are **0**'s exclusively.

The following routine (in MATLAB code) verifies the orthonormality of c(:,k) set.

```
i = 0;
for n = 1:q-1
  for m = n+1:q
    i = i+1;
    y(:,i) = c(:,n)&c(:,m);
  end
end
```

**Simulation of BCMM**

The simulations demonstrate that the BOP algorithm improves the storage capacity of the BAM to the maximum. Stated another way, the largest number of patterns that can be reliably stored is reached.

### III. CONCLUSIONS

A Boolean version of the Correlation Matrix Memory (CMM) was introduced, which is very useful to work with binary memories. A novel Boolean Orthonormalization Process (BOP) was presented to convert a non-orthonormal Boolean basis, i.e., a set of non-orthogonal binary vectors (in a Boolean sense) to an orthonormal Boolean basis, i.e., a set of orthonormal binary vectors (in a Boolean sense). The simulation results shows that it is possible to improve the performance of Boolean CMM thanks BOP algorithm. Besides, the BOP algorithm has a lot of additional fields of applications, e.g.: Compression of palettized images by color, Steganography, Bi-level image processing, among others image processing appli-cations, see [12-17], besides, it is very useful in Hopfield Networks [1, 18-23], etc.

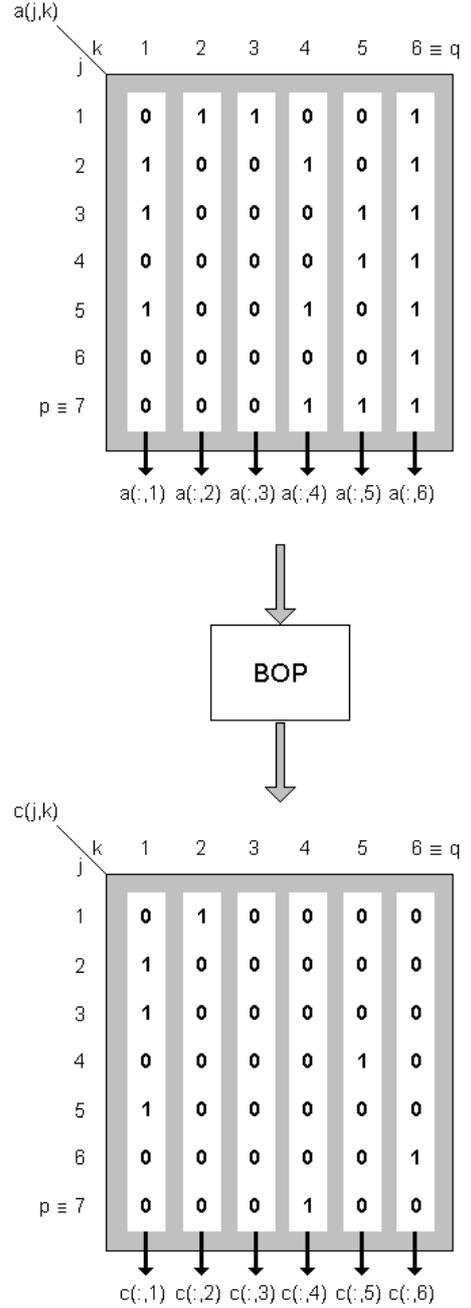

Fig. 4 Example of BOP algorithm application in MATLAB syntax

It is important to mention that BOP algorithm should be applied element-to-element in the implicated vectors, see

Eq.10, which produces the desired orthonormality effect *in a Boolean sense*.

Finally, the BOP demonstrated to be an extremely stable and fast algorithm.